# An Enhanced Hybrid MobileNet

Hong-Yen Chen
Department of Electrical Engineering
National Taiwan Normal University
Taipei, Taiwan
40475009H@ntnu.edu.tw

Chung-Yen Su*
Department of Electrical Engineering
National Taiwan Normal University
Taipei, Taiwan
scy@ntnu.edu.tw

*Abstract*—Complicated and deep neural network models can achieve high accuracy for image recognition. However, they require a huge amount of computations and model parameters, which are not suitable for mobile and embedded devices. Therefore, MobileNet was proposed, which can reduce the number of parameters and computational cost dramatically. The main idea of MobileNet is to use a depthwise separable convolution. Two hyper-parameters, a width multiplier and a resolution multiplier are used to the trade-off between the accuracy and the latency. In this paper, we propose a new architecture to improve the MobileNet. Instead of using the resolution multiplier, we use a depth multiplier and combine with either Fractional Max Pooling or the max pooling. Experimental results on CIFAR database show that the proposed architecture can reduce the amount of computational cost and increase the accuracy simultaneously[1].

*Keywords—deep learning; MobileNet; neural networks; image classifier; image recognition*

## I. INTRODUCTION

Since AlexNet [1] achieved ImageNet's Champion [2], the general trend of deep learning is to make more complicated and deeper networks to get a higher accuracy [3-6]. However, complicated networks cost lots of resources and are not suitable for mobile and embedded devices. To solve the problems, models with less parameters are getting more and more attention [7-10]. MobileNet [11] is one of them, and it provides a solution for mobile and embedded devices.

Instead of using the standard convolution, MobileNet uses a special convolution called depthwise separable convolution. With the depthwise separable convolution, it needs about one-eighth of computational cost and has only a little drop in accuracy. For the trade-offs of computational cost, size of model parameters, and accuracy, MobileNet also provides two hyper-parameters. Those are α, named width multiplier, and ρ, named resolution multiplier. Although adjusting α or ρ can reduce the computational cost or parameter size of models tremendously, it always causes accuracy to drop.

In this paper, we present a new model architecture to solve this problem. With the proposed architecture, we are able to increase the accuracy and reduce the computational cost simultaneously.

The paper is organized as follows. Section II introduces the prior works about the model structure of MobileNet and various pooling patterns. Section III describes the main idea of the proposed architecture about using a new hyper-parameter, depth multiplier, and adding two different kinds of max pooling to MobileNet. Section IV shows experimental results on CIFAR-10 and CIFAR-100 databases. Section V gives a concluding remark.

## II. PRIOR WORKS

Depthwise separable convolution, which is a form of factorized convolution, is the core layer in MobileNet, and its characteristic is to split the standard convolution into a depthwise convolution and a 1 × 1 convolution named pointwise convolution. In the MobileNet, the depthwise convolution means that each of filters only uses a single input channel to do convolution. Then the pointwise convolution integrates the depthwise convolution output and performs a 1 × 1 convolution. Depthwise separable convolution uses between 8 to 9 times less computational cost than the standard convolution. Although the computation has already been reduced by using depthwise separable convolution, a different approach to reduce computation is shrinking or factorizing. In MobileNet, two hyper-parameters are also presented for shrinking and factorizing, which are α named width multiplier and ρ named resolution multiplier. Both the values of α and ρ are in the range (0, 1]. Reducing the value of α corresponds to reducing the number of filters, while reducing the value of ρ corresponds to reducing the image resolution. By adjusting both the values of α and ρ, we can obtain a small network and reduce computation. However, it always goes with reduced accuracy.

There are various kinds of pooling used in neural network models. Max Pooling Kernel size 2 × 2 Stride 2 is the common choice for building convolutional networks because it keeps most of the features and quickly reduces the size of the hidden layers. But according to [1], the effect of $3 \times 3$ pooling regions overlapping with stride 2 is greater than the common pooling. Different from the general pooling methods that downsample the images in integer multiples, Fractional Max Pooling [12] is able to downsample the images in decimals and able to increase the accuracy of networks.

---

[1] This work is partly supported by Ministry of Science and Technology, R.O.C. under Contract No. MOST 106-2221-E-003-011.

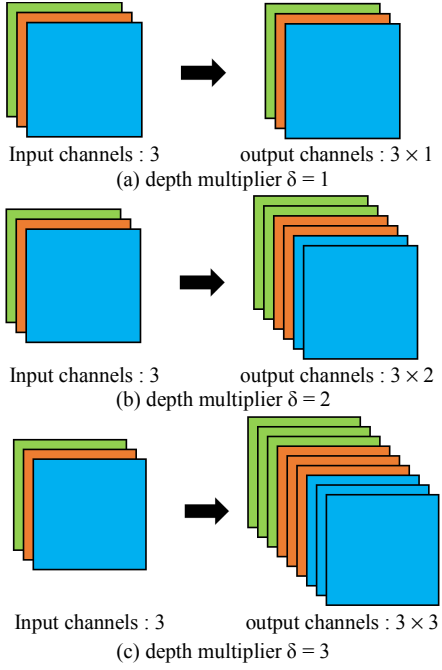

Figure 1. The operating principles of depth multiplier δ.

**Table 1. MobileNet in different α and δ architecture**

| Type / Stride | Filter Shape | Input Size |
|---|---|---|
| Conv / s2 | $3 \times 3 \times 3 \times 32\alpha$ | $32 \times 32 \times 3$ |
| Conv dw / s1 | $3 \times 3 \times (32\alpha \times \delta)$ | $16 \times 16 \times 32\alpha$ |
| Conv / s1 | $1 \times 1 \times (32\alpha \times \delta) \times 64\alpha$ | $16 \times 16 \times (32\alpha \times \delta)$ |
| Conv dw / s2 | $3 \times 3 \times (64\alpha \times \delta)$ | $16 \times 16 \times 64\alpha$ |
| Conv / s1 | $1 \times 1 \times (64\alpha \times \delta) \times 128\alpha$ | $8 \times 8 \times (64\alpha \times \delta)$ |
| Conv dw / s1 | $3 \times 3 \times (128\alpha \times \delta)$ | $8 \times 8 \times 128\alpha$ |
| Conv / s1 | $1 \times 1 \times (128\alpha \times \delta) \times 128\alpha$ | $8 \times 8 \times (128\alpha \times \delta)$ |
| Conv dw / s2 | $3 \times 3 \times (128\alpha \times \delta)$ | $8 \times 8 \times 128\alpha$ |
| Conv / s1 | $1 \times 1 \times (128\alpha \times \delta) \times 256\alpha$ | $4 \times 4 \times (128\alpha \times \delta)$ |
| Conv dw / s1 | $3 \times 3 \times (256\alpha \times \delta)$ | $4 \times 4 \times 256\alpha$ |
| Conv / s1 | $1 \times 1 \times (256\alpha \times \delta) \times 256\alpha$ | $4 \times 4 \times (256\alpha \times \delta)$ |
| Conv dw / s2 | $3 \times 3 \times (256\alpha \times \delta)$ | $4 \times 4 \times 256\alpha$ |
| Conv / s1 | $1 \times 1 \times (256\alpha \times \delta) \times 512\alpha$ | $2 \times 2 \times (256\alpha \times \delta)$ |
| 5× Conv dw / s1 Conv / s1 | $3 \times 3 \times (512\alpha \times \delta)$ $1 \times 1 \times (512\alpha \times \delta) \times 512\alpha$ | $2 \times 2 \times 512\alpha$ $2 \times 2 \times (512\alpha \times \delta)$ |
| Conv dw / s2 | $3 \times 3 \times (512\alpha \times \delta)$ | $2 \times 2 \times 512\alpha$ |
| Conv / s1 | $1 \times 1 \times (512\alpha \times \delta) \times 1024\alpha$ | $1 \times 1 \times (512\alpha \times \delta)$ |
| Conv dw / s1 | $3 \times 3 \times (1024\alpha \times \delta)$ | $1 \times 1 \times 1024\alpha$ |
| Conv / s1 | $1 \times 1 \times (1024\alpha \times \delta) \times 1024\alpha$ | $1 \times 1 \times (1024\alpha \times \delta)$ |
| Avg Pool / s1 | Global average pooling | $1 \times 1 \times 1024\alpha$ |
| FC / s1 | $1 \times 1 \times 1024\alpha \times$ classnum | $1 \times 1 \times 1024\alpha$ |
| Softmax / s1 | Classifier | $1 \times 1 \times$ classnum |

### III. THE PROPOSED METHODS

In this section, we present a new hyper-parameter δ named depth multiplier and describe how to use it to MobileNet.

#### A. MobileNet architecture in different α and δ

Our main idea comes from that one input channel can result in multiple features after the depthwise convolution. To do so, we introduce a new hyper-parameter δ named depth multiplier in the depthwise convolution to change the number of feature maps corresponding to each input channel. The parameter δ is an integer, typically values are 1, 2, and 4. δ = 2 means that two feature maps are obtained for each input channel after the depthwise convolution. For the images with 3 input channels, the number of feature maps becomes 3δ. The operating principles of δ are illustrated in Figure 1. Note that δ = 1 is the baseline in MobileNet. If δ > 1, both the computational cost and the size of model parameters will be increased. To reduce the overall computational cost and the size of parameters, we need the same width multiplier α as that in MobileNet. The role of the width multiplier α is to thin a network uniformly at each layer. By adjusting the values of δ and α adequately, we can reduce the overall computational cost and the size of parameters in the MobileNet. **Table 1** shows the proposed architecture. Note that we do not use the resolution parameter ρ herein. This is because we found that for the applications of object detection, reducing the image resolution may drastically lower the detection rate.

Let $D_K \times D_K$ be the size of convolution kernel $K$, $M$ be the number of input channels, and $N$ be the number of output channels. We have the number of parameters of a depthwise convolution as:

$$D_K \cdot D_K \cdot (\alpha M \cdot \delta) \quad (1)$$

and the number of parameters of a pointwise convolution as:

$$(\alpha M \cdot \delta) \cdot \alpha N \quad (2)$$

Let $D_F \times D_F$ be the size of feature map. A depthwise convolution has the computational cost:

$$D_K \cdot D_K \cdot (\alpha M \cdot \delta) \cdot D_F \cdot D_F \quad (3)$$

A pointwise convolution has the computational cost:

$$(\alpha M \cdot \delta) \cdot \alpha N \cdot D_F \cdot D_F \quad (4)$$

We can calculate both the ratios of the number of parameters and computational cost between the modified depthwise separable convolution and the original depthwise separable convolution, i.e. α = 1 and δ = 1.

The ratio of number of parameters is given by :

$$\frac{D_K \cdot D_K \cdot (\alpha M \cdot \delta) + (\alpha M \cdot \delta) \cdot \alpha N}{D_K \cdot D_K \cdot M + M \cdot N}$$

Table 2. MobileNet combined with Max Pooling Kernel size 3 × 3 Stride 2 in different α and δ architecture

| Type / Stride | Filter Shape | Input Size |
|---|---|---|
| Conv / s1 | 3 × 3 × 3 × 32α | 32 × 32 × 3 |
| Maxpool / s2 | Max pooling kernel 3×3 stride2 | 32 × 32 × 32α |
| Conv dw / s1 | 3 × 3 × (32α× δ) | 16 × 16 × 32α |
| Conv / s1 | 1 × 1 × (32α× δ) × 64α | 16 × 16 × (32α× δ) |
| Maxpool / s2 | Max pooling kernel 3×3 stride2 | 16 × 16 × 64α |
| Conv dw / s1 | 3 × 3 × (64α× δ) | 8 × 8 × 64α |
| Conv / s1 | 1 × 1 × (64α× δ) × 128α | 8 × 8 × (64α× δ) |
| Conv dw / s1 | 3 × 3 × (128α× δ) | 8 × 8 × 128α |
| Conv / s1 | 1 × 1 × (128α× δ) × 128α | 8 × 8 × (128α× δ) |
| Maxpool / s2 | Max pooling kernel 3×3 stride2 | 8 × 8 × 128α |
| Conv dw / s1 | 3 × 3 × (128α× δ) | 4 × 4 × 128α |
| Conv / s1 | 1 × 1 × (128α× δ) × 256α | 4 × 4 × (128α× δ) |
| Conv dw / s1 | 3 × 3 × (256α× δ) | 4 × 4 × 256α |
| Conv / s1 | 1 × 1 × (256α× δ) × 256α | 4 × 4 × (256α× δ) |
| Maxpool / s2 | Max pooling kernel 3×3 stride2 | 4 × 4 × 256α |
| Conv dw / s1 | 3 × 3 × (256α× δ) | 2 × 2 × 256α |
| Conv / s1 | 1 × 1 × (256α× δ) × 512α | 2 × 2 × (256α× δ) |
| 5 × Conv dw / s1<br>Conv / s1 | 3 × 3 × (512α× δ)<br>1 × 1 × (512α× δ) × 512α | 2 × 2 × 512α<br>2 × 2 × (512α× δ) |
| Maxpool / s2 | Max pooling kernel 3×3 stride2 | 2 × 2 × 512α |
| Conv dw / s1 | 3 × 3 × (512α× δ) | 1 × 1 × 512α |
| Conv / s1 | 1 × 1 × (512α× δ) × 1024α | 1 × 1 × (512α× δ) |
| Conv dw / s1 | 3 × 3 × (1024α× δ) | 1 × 1 × 1024α |
| Conv / s1 | 1 × 1 × (1024α× δ) × 1024α | 1 × 1 × (1024α× δ) |
| Avg pool / s1 | Global average pooling | 1 × 1 × 1024α |
| FC / s1 | 1 × 1 × 1024α× classnum | 1 × 1 × 1024α |
| Softmax / s1 | Classifier | 1 × 1 × classnum |

$$= \frac{(\alpha D_K^2 + \alpha^2 N) \cdot \delta}{D_K^2 + N} \cong \alpha^2 \cdot \delta$$

The ratio of computational cost is:

$$\frac{D_K \cdot D_K \cdot (\alpha M \cdot \delta) \cdot D_F \cdot D_F + (\alpha M \cdot \delta) \cdot \alpha N \cdot D_F \cdot D_F}{D_K \cdot D_K \cdot M \cdot D_F \cdot D_F + M \cdot N \cdot D_F \cdot D_F}$$

$$= \frac{(\alpha D_K^2 + \alpha^2 N) \cdot \delta}{D_K^2 + N} \cong \alpha^2 \cdot \delta$$

For example, assuming that $N = 128$, $M = 64$, $D_K = 3$, $D_F = 32$, $\alpha = 0.5$ and $\delta = 2$, we can obtain that both the ratio of number of parameters and the ratio of computational cost are 0.53, meaning that near half of the number of parameters and half of the computational cost can be saved.

*B. MobileNet combined with Max Pooling Kernel size 3×3 Stride 2 in different α and δ architecture*

MobileNet does not use any pooling. However, a pooling can keep more features than only using the stride in the convolution layer. Therefore, we try to add a pooling to the proposed architecture and expect that it may increase accuracy. We replace the original stride in depthwise separable convolution to Max Pooling Kernel size 3 × 3 Stride 2. The modified architecture is shown in **Table 2.**

Table 3. MobileNet combined with Fractional Max Pooling Stride 1.4 in different α and δ architecture

| Type / Stride | Filter Shape | Input Size |
|---|---|---|
| Conv / s1 | 3 × 3 × 3 × 32α | 32 × 32 × 3 |
| FMP / s1.4 | Fractional Max Pooling s1.4 | 32 × 32 × 32α |
| Conv dw / s1 | 3 × 3 × (32α× δ) | 22 × 22 × 32α |
| Conv / s1 | 1 × 1 × (32α× δ) × 64α | 22 × 22 × (32α× δ) |
| Conv dw / s1 | 3 × 3 × (64α× δ) | 22 × 22 × 64α |
| Conv / s1 | 1 × 1 × (64α× δ) × 128α | 22 × 22 × (64α× δ) |
| FMP / s1.4 | Fractional Max Pooling s1.4 | 22 × 22 × 128α |
| Conv dw / s1 | 3 × 3 × (128α× δ) | 15 × 15 × 128α |
| Conv / s1 | 1 × 1 × (128α× δ) × 128α | 15 × 15 × (128α× δ) |
| Conv dw / s1 | 3 × 3 × (128α× δ) | 15 × 15 × 128α |
| Conv / s1 | 1 × 1 × (128α× δ) × 256α | 15 × 15 × (128α× δ) |
| FMP / s1.4 | Fractional Max Pooling s1.4 | 15 × 15 × 256α |
| Conv dw / s1 | 3 × 3 × (256α× δ) | 10 × 10 × 256α |
| Conv / s1 | 1 × 1 × (256α× δ) × 256α | 10 × 10 × (128α× δ) |
| Conv dw / s1 | 3 × 3 × (256α× δ) | 10 × 10 × 256α |
| Conv / s1 | 1 × 1 × (256α× δ) × 512α | 10 × 10 × (256α× δ) |
| FMP / s1.4 | Fractional Max Pooling s1.4 | 10 × 10 × 512α |
| 2 × Conv dw / s1<br>Conv / s1 | 3 × 3 × (512α× δ)<br>1 × 1 × (512α× δ) × 512α | 6 × 6 × 512α<br>6 × 6 × (512α× δ) |
| FMP / s1.4 | Fractional Max Pooling s1.4 | 6 × 6 × 512α |
| 2 × Conv dw / s1<br>Conv / s1 | 3 × 3 × (512α× δ)<br>1 × 1 × (512α× δ) × 512α | 4 × 4 × 512α<br>4 × 4 × (512α× δ) |
| FMP / s1.4 | Fractional Max Pooling s1.4 | 4 × 4 × 512α |
| Conv dw / s1 | 3 × 3 × (512α× δ) | 2 × 2 × 512α |
| Conv / s1 | 1 × 1 × (512α× δ) × 512α | 2 × 2 × (512α× δ) |
| Conv dw / s1 | 3 × 3 × (512α× δ) | 2 × 2 × 512α |
| Conv / s1 | 1 × 1 × (512α× δ) × 1024α | 2 × 2 × (512α× δ) |
| FMP / s1.4 | Fractional Max Pooling s1.4 | 2 × 2 × 1024α |
| Conv dw / s1 | 3 × 3 × (1024α× δ) | 1 × 1 × 1024α |
| Conv / s1 | 1 × 1 × (1024α× δ) × 1024α | 1 × 1 × (1024α× δ) |
| Avg Pool / s1 | Global average pooling | 1 × 1 × 1024α |
| FC / s1 | 1 × 1 × 1024α× classnum | 1 × 1 × 1024α |
| Softmax / s1 | Classifier | 1 × 1 × classnum |

*C. MobileNet combined with Fractional Max Pooling Stride 1.4 in different α and δ architecture*

Since the Fractional Max Pooling can keep bigger feature maps than the general pooling, we also try to use it. We remove all strides in the depthwise separable convolution and add the Fractional Max Pooling Stride 1.4 to the proposed architecture. The new model architecture is shown in **Table 3**.

IV. EXPERIMENTS

In this section, we will show the image classification results on the architectures of **Table 1**, **Table 2**, and **Table 3**.

All network models are trained in Keras [13] with a backend Tensorflow [14]. We use the optimizer named Adam [15] with 10% dropout in a fully connection layer. The databases in the experiments are CIFAR-10 consisting of images drawn from 10 classes and CIFAR-100 from 100 classes. The training and test sets of CIFAR-10 and CIFAR-100 respectively contain 50,000 and 10,000 images without any data augmentation.

### A. MobileNet in different α and δ

In this experiment, we use different values of α and δ to the proposed architecture in **Table 1**. **Table 4** tabulates the results of accuracy, computational cost and number of parameters. As we can see in **Table 4**, accuracy increases 1.9% for CIFAR-10 and 4.4% for CIFAR-100 when δ = 4 and α = 0.5. In the settings, the computational cost only increases 2.1% for CIFAR-10 and 1.4% for CIFAR-100, while the number of parameters decreases only 1.5% for CIFAR-10 and 0% for CIFAR-100.

### B. MobileNet combined with Max Pooling Kernel size 3×3 Stride 2 in different α and δ

In this experiment, we use different values of α and δ to the proposed architecture in **Table 2**. **Table 5** tabulates the results of accuracy, computational cost and number of parameters. In **Table 5** we find that using the proposed architecture with Max Pooling Kernel size 3 × 3 Stride 2 compared to the baseline (MobileNet) can increase accuracy and reduce both computational cost and parameters at the same time. For example, using the settings, δ = 2 and α = 0.5, increases accuracy by 0.6% for CIFAR-10. Meanwhile, it reduces computational cost by 47% and parameters by 49%. As for CIFAR-100, using the same setting increases accuracy by 1.9% and reduces both computational cost by 48% and parameters by 49%.

### C. MobileNet combined with Fractional Max Pooling Stride 1.4 in different α and δ

In this experiment, we use different values of α and δ to the proposed architecture in **Table 3**. The results are shown in **Table 6.** From this table, we see that using the proposed architecture with Fractional Max Pooling Stride 1.4 compared to the baseline can increase accuracy but largely reduce parameters. For example, setting δ = 2 and α = 0.25, we can increase accuracy by 5.4% for CIFAR-10 and by 11.7% for CIFAR-100. Meanwhile, we can reduce computational cost by 18% and parameters by 87% for both CIFAR-10 and CIFAR-100.

## V. Conclusion

We proposed a new model architecture to improve the MobileNet. We introduced a new depth multiplier to increase the number of feature maps corresponding to the input image channels. Meanwhile, we used the width multiplier to balance the accuracy and computational cost. We then demonstrated how to modify the architecture with either the Max Pooling Kernel size 3 × 3 Stride 2 or the Fractional Max Pooling Stride 1.4. Compared with the original MobileNet, the modified architectures not only can increase accuracy but also can reduce both computational cost and parameters.

Table 4. Results on MobileNet in different α and δ

| Model MobileNet | CIFAR-10 Accuracy | Million Mult-Adds | Million Parameters |
|---|---|---|---|
| **δ = 1,α = 1 Baseline** | **76.7%** | **11.58** | **3.23** |
| **δ = 2,α = 1** | **81.1%** | **22.96** | **6.44** |
| **δ = 2,α = 0.5** | **74.9%** | **5.97** | **1.65** |
| **δ = 2,α = 0.25** | **70.2%** | **1.61** | **0.43** |
| **δ = 4,α = 1** | **81.7%** | **45.69** | **12.85** |
| **δ = 4,α = 0.5** | **78.6%** | **11.83** | **3.28** |
| **δ = 4,α = 0.25** | **73.8%** | **3.16** | **0.85** |
| Model MobileNet | CIFAR-100 Accuracy | Million Mult-Adds | Million Parameters |
| **δ = 1,α = 1 Baseline** | **39.1%** | **11.68** | **3.33** |
| **δ = 2,α = 1** | **41.8%** | **22.97** | **6.53** |
| **δ = 2,α = 0.5** | **38.2%** | **5.98** | **1.70** |
| **δ = 2,α = 0.25** | **37.1%** | **1.62** | **0.46** |
| **δ = 4,α = 1** | **44.3%** | **45.70** | **12.94** |
| **δ = 4,α = 0.5** | **43.5%** | **11.84** | **3.33** |
| **δ = 4,α = 0.25** | **37.9%** | **3.17** | **0.88** |

Table 5. Results on MobileNet combined with Max Pooling Kernel size 3 × 3 Stride 2 in different α and δ

| Model MobileNet+MP3 | CIFAR-10 Accuracy | Million Mult-Adds | Million Parameters |
|---|---|---|---|
| **δ = 2,α = 1** | **81.3%** | **23.18** | **6.44** |
| **δ = 2,α = 0.5** | **77.3%** | **6.08** | **1.65** |
| **δ = 2,α = 0.25** | **71.2%** | **1.67** | **0.43** |
| **δ = 4,α = 1** | **82.3%** | **45.91** | **12.85** |
| **δ = 4,α = 0.5** | **79.7%** | **11.94** | **3.28** |
| **δ = 4,α = 0.25** | **74.8%** | **3.22** | **0.85** |
| **Baseline MobileNet** | **76.7%** | **11.58** | **3.23** |
| Model MobileNet+MP3 | CIFAR-100 Accuracy | Million Mult-Adds | Million Parameters |
| **δ = 2,α = 1** | **42.3%** | **23.28** | **6.53** |
| **δ = 2,α = 0.5** | **41.0%** | **6.13** | **1.70** |
| **δ = 2,α = 0.25** | **38.1%** | **1.64** | **0.46** |
| **δ = 4,α = 1** | **46.2%** | **46.01** | **12.94** |
| **δ = 4,α = 0.5** | **43.5%** | **12.0** | **3.33** |
| **δ = 4,α = 0.25** | **37.9%** | **3.24** | **0.88** |
| **Baseline MobileNet** | **39.1%** | **11.68** | **3.33** |

Table 6. Results on MobileNet combined with Fractional Max Pooling Stride 1.4 in different α and δ

| Model MobileNet+FMP | CIFAR-10 Accuracy | Million Mult-Adds | Million Parameters |
|---|---|---|---|
| **δ = 2,α = 1** | **88.9%** | **139** | **6.44** |
| **δ = 2,α = 0.5** | **86.1%** | **35.9** | **1.65** |
| **δ = 2,α = 0.25** | **82.1%** | **9.5** | **0.43** |
| **δ = 4,α = 1** | **89.6%** | **277** | **12.85** |
| **δ = 4,α = 0.5** | **87.8%** | **71.2** | **3.28** |
| **δ = 4,α = 0.25** | **84.5%** | **18.9** | **0.85** |
| **Baseline MobileNet** | **76.7%** | **11.58** | **3.23** |
| Model MobileNet+FMP | CIFAR-100 Accuracy | Million Mult-Adds | Million Parameters |
| **δ = 2,α = 1** | **59.3%** | **139** | **6.53** |
| **δ = 2,α = 0.5** | **56.9%** | **36** | **1.70** |
| **δ = 2,α = 0.25** | **50.8%** | **9.6** | **0.46** |
| **δ = 4,α = 1** | **60.9%** | **277** | **12.94** |
| **δ = 4,α = 0.5** | **58.6%** | **71.3** | **3.33** |
| **δ = 4,α = 0.25** | **54.1%** | **18.9** | **0.88** |
| **Baseline MobileNet** | **39.1%** | **11.68** | **3.33** |

## REFERENCES

[1] A. Krizhevsky, I. Sutskever, and G. E. Hinton. Imagenet Classification with deep convolutional neural networks. In Advances in neural information processing systems, pages 1097–1105, 2012.

[2] O. Russakovsky, J. Deng, H. Su, J. Krause, S. Satheesh, S. Ma, Z. Huang, A. Karpathy, A. Khosla, M. Bernstein, *et al*. Imagenet large scale visual recognition challenge International Journal of Computer Vision, 115(3):211-252, 2015.

[3] K. Simonyan and A. Zisserman. Very deep convolutional networks for large-scale image recognitions. *arXiv preprint arXiv:1409.1556,* 2014.

[4] C. Szegedy, V. Vanhoucke, S. Ioffe, J. Shlens, and Z. Wojna. Rethinking the inception architecture for computer vision. *arXiv preprint arXiv: 1512.00567,* 2015.

[5] K. He, X. Zhang, S. Ren, and J. Sun. Deep residual learning for image recognition. *arXiv preprint arXiv :1512.03385,* 2015.

[6] G. Huang, Z. Liu, K. Q. Weinberger and L. Maaten. Densely connected convolutional networks. *arXiv preprint arXiv:1608.06993,*2016.

[7] J. Jin, A. Dundar,.and E. Culurciello. Flattened convolutional neural networks for feedforward acceleration. *arXiv preprint arXiv: 1412. 5474,* 2014.

[8] F. N. Iandola, S. Han, M. W. Moskewicz, K. Ashraf, W. J. Dally and K. Keutzer. SqueezeNet : AlexNet-level accuracy with 50x fewer parameters and < 0.5 MB model size. *arXiv preprint arXiv:1602. 07360,* 2016.

[9] M. Rastegari, V. Ordonez, J. Redmon, and A. Farhadi. Xnornet : Imagenet classification using binary convolutional neural networks. a*rXiv preprint arXiv:1603.05279,* 2016.

[10] M. Wang, B. Liu, and H. Foroosh. Factorized convolutional neural networks. *arXiv preprint arXiv: 1608.04337,* 2016.

[11] A. G. Howard, M. Zhu,. B. Chen, D. Kalenichenko, W. Wang, T. Weyand, M. Andreetto and H. Adam. MobileNet : Efficient Convolutional Neural Networks for Mobile Applications. *arXiv preprint arXiv:1704.04861,*2017.

[12] B. Graham. Fractional Max-Pooling. *arXiv preprint arXiv: 1412.6071,* 2014.

[13] @misc{chollet2015keras, title={Keras}, author={Chollet, ran\c{c}ois and others}, year={2015}, publisher={GitHub}, howpublished ={\url{https://github.com/fchollet/keras}}, }

[14] M. Abadi, A. Agarwal, P. Barham, E. Brevdo, Z. Chen, C. Citro, G. S. Corrado, A. Davis, J. Dean, M.Devin, et al. Tensorflow:Large-scale machine learning onheterogeneous systems, 2015. Software available from tensorflow org, 1, 2015.

[15] D. P. Kingma and J. Ba. Adam: A Method for Stochastic Optimization. *arXiv preprint arXiv: 1412.6980,* 2014.